\documentclass[letterpaper, 10 pt, conference]{ieeeconf} 
\IEEEoverridecommandlockouts % This command is only
\usepackage{tabularx, graphicx,
  amsmath,amsfonts,siunitx,pifont,amssymb,subcaption}
\usepackage{xcolor, bm, float, multirow, multicol, courier, mathtools, url}
\usepackage[linesnumbered,ruled,vlined]{algorithm2e} \pdfminorversion=4
\let\labelindent\relax \usepackage{enumitem} \usepackage{graphicx}

\DeclareMathOperator*{\argmin}{argmin}
\DeclareMathOperator*{\argmax}{argmax}

\usepackage{siunitx}

\usepackage{pdfpages}
\usepackage{cite}
\usepackage{algpseudocode}
\usepackage{float}
\usepackage{colortbl}
\usepackage{dblfloatfix}    % To enable figures at the bottom of page

\def\secref#1{Sec.~\ref{#1}}
\def\figref#1{Fig.~\ref{#1}}
\def\tabref#1{Tab.~\ref{#1}}
\def\eqref#1{Eq.~(\ref{#1})}

\title{\LARGE \bf Online LiDAR-SLAM for Legged Robots\\ with Robust Registration and Deep-Learned Loop Closure}

\author{Milad Ramezani, Georgi Tinchev, Egor Iuganov and Maurice Fallon% <-this % stops a space
	\thanks{This research was supported by the
Innovate UK-funded ORCA Robotics Hub (EP/R026173/1) and the EU H2020 Project MEMMO. M. Fallon is supported by a Royal Society University Research Fellowship.
	}
	\thanks{The authors are with the Oxford Robotics Institute, University of Oxford, UK.
		{\tt\small \{milad, gtinchev, mfallon\}@robots.ox.ac.uk, egor.iuganov@wadham.ox.ac.uk}}%
}

\begin{document}
	
\setlength{\abovedisplayskip}{4pt}
\setlength{\belowdisplayskip}{4pt}

\maketitle \thispagestyle{empty} \pagestyle{empty}

%%%%%%%%%%%%%%%%%%%%%%%%%%%%%%%%%%%%%%%%%%%%%%%%%%%%%%%%%%%%%%%%%%%%%%%%%%%%%%%%
\begin{abstract}
%%1. what is the topic?  2. state the
%%problem you tackle 3. summarize why
%%it has not been done properly before
%%4. how did we tackle it? what is the
%%idea 5. how did we implement it
%%6. what is the impact	
In this paper, we present a factor-graph LiDAR-SLAM system which incorporates
a state-of-the-art deeply learned feature-based loop closure detector to enable a
legged robot to localize and map in industrial environments.
These facilities can be badly lit and comprised of indistinct metallic structures,
thus our system uses only LiDAR sensing and was developed to run on the quadruped robot's navigation PC.
Point clouds are accumulated using an inertial-kinematic 
state estimator before being aligned using ICP registration.
To close loops we use a loop proposal mechanism which matches individual \textit{segments} between clouds.
We trained a descriptor offline to match these segments. The efficiency of our method comes 
from carefully designing the network architecture to minimize the number of parameters such
that this deep learning method can be deployed in real-time using only the CPU of a legged robot,
a major contribution of this work. The set of odometry and loop closure factors are updated 
using pose graph optimization. Finally we present an efficient risk alignment prediction method which 
verifies the reliability of the registrations. Experimental results at an industrial facility demonstrated the 
robustness and flexibility of our system, including autonomous following paths derived from the SLAM 
map.
\end{abstract}

%\vspace{-0.4em}
\section{Introduction}
\vspace{-0.5em}
Robotic mapping and localization have been heavily studied over the last two decades and provide the 
perceptual basis for many different tasks such as motion planning, control and manipulation. A vast 
body of research has been carried out to allow a robot to determine where it is located in an unknown 
environment, to navigate and to accomplish tasks robustly online. Despite substantial progress, enabling an 
autonomous mobile robot to operate robustly for long time periods in 
complex environments, is still an active research area.

Visual SLAM has shown substantial progress \cite{mur2015orb,engel2014lsd}, with much
work focusing on overcoming the challenge of changing lighting variations \cite{porav2018adversarial}. 
Instead, in this work we focus on LiDAR as our primary sensor modality. Laser measurements are 
actively illuminated and precisely sense the environment at long ranges which is attractive for 
accurate motion estimation and mapping. In this work, we focus on the LiDAR-SLAM specifically 
for legged robots. 

\begin{figure}
\includegraphics[width=\linewidth]{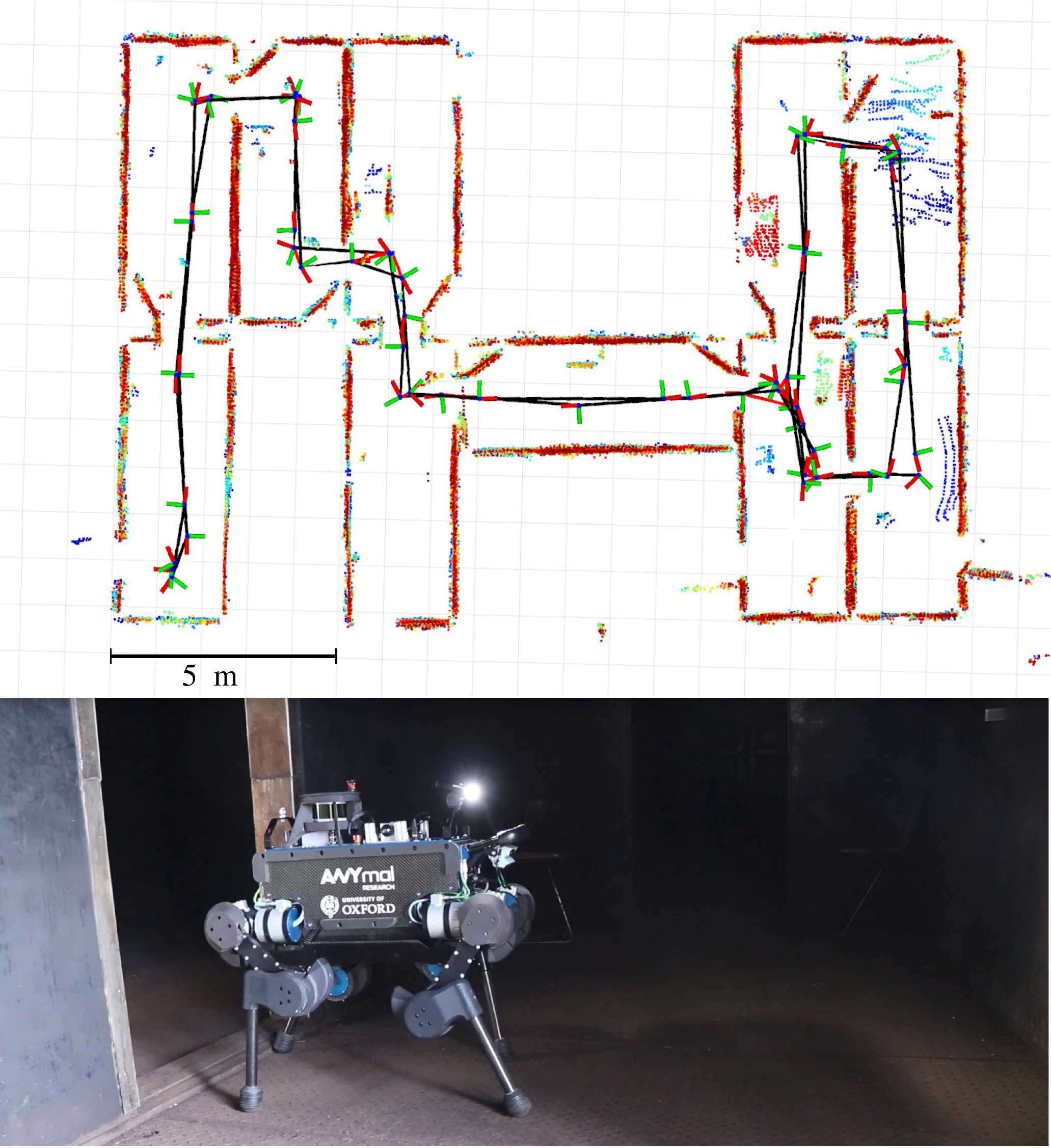}
\caption{\small{Bird's-eye view of a map constructed by the ANYmal quadruped exploring an unlit, windowless, 
industrial facility with our proposed LiDAR-SLAM system. The approach can rapidly detect and verify loop-closures (red lines) so
as to construct an accurate map. The approach uses only LiDAR due to the difficult
illumination conditions negating the use of visual approaches.}}
\label{fig:teaser}
\vspace{-2em}
\end{figure}

The core odometry of our SLAM system is based on Iterative Closest Point (ICP) registration of 3D point
clouds from a LiDAR (Velodyne) sensor. Our approach builds upon our previous work of Autotuning ICP (AICP), 
originally proposed in~\cite{Nobili17icra}, which analyzes the content of incoming clouds to 
robustify registration.
Initialization of AICP is provided by the robot's state estimator\cite{bloesch2013state}, 
which estimates the robot pose by fusing the inertial measurements with the robot's kinematics and joint 
information in a recursive approach. Our LiDAR odometry has drift below
1.5$\%$ of distance traveled.

The contributions of our paper are summarized as follows:

\begin{enumerate}
	\item A LiDAR-SLAM system for legged robots based on ICP registration. The approach uses 
the robot's odometry for accurate initialization and a factor graph for 
global optimization using GTSAM \cite{dellaert2017gtsam}.
	\item A verification metric which quantifies the reliability of 
point cloud registration. Inspired by the alignment risk prediction method developed by Nobili~\textit{et al.}~\cite{Nobili18icra}, we 
propose a verification system based on efficient k-d tree search \cite{bentley1975multidimensional} to 
confirm whether a loop closure is well constrained. In contrast to~\cite{Nobili18icra}, our modified 
verification method is suitable for online operation.
	\item A demonstration of feature-based loop-closure 
detection which takes advantage of deep learning, specifically designed to be deployable on a mobile CPU 
during run time. It shows substantially lower computation times when compared to a 
geometric loop closure method, that scales with the number of surrounding clouds.
	\item A real-time demonstration of the algorithm on the ANYmal quadruped robot. 
We show how the robot can use the map representation to 
plan safe routes and to autonomously return home on request. We provide
detailed quantitative analysis against ground truth maps.
\end{enumerate}

The remainder of this paper is structured as follows: \secref{sec:relatedWorks} presents 
related works followed by a description of the platform and experimental scenario in 
\secref{sec:scenario}.~\secref{sec:approach} details different components of our SLAM 
system.~\secref{sec:exp-eval} presents evaluation studies before a conclusion and future works are 
drawn in~\secref{sec:conclusion}.
%\vspace{-0.4em}
\section{Related Works}
\vspace{-0.2em}
\label{sec:relatedWorks}
This section provides a literature review of perception in walking robots and LiDAR-SLAM systems in 
general.

%\vspace{-0.5em}
\subsection{Perception systems on walking robots}
%\vspace{-0.5em}
Simultaneous Localization and Mapping (SLAM) is a key capability for walking robots and 
consequently their autonomy. Example systems include, the mono visual SLAM system of~\cite{stasse2006real}
which ran on the HRP-2 humanoid robot and is based on an Extended Kalman Filter (EKF) 
framework. 
In contrast,~\cite{kwak20093d} leveraged a particle filter to estimate the posterior of their SLAM method, again on HRP-2. 
Oriolo~\textit{et al.} \cite{oriolo2012vision} demonstrated visual odometry on the Nano bipedal robot by 
tightly coupling visual information with the robot's kinematics and inertial sensing within an EKF. 
However, these approaches were reported to acquire only a sparse map of the environment 
on the fly, substantially limiting the robot's perception.

Ahn~\textit{et al.} \cite{ahn2012board} presented a vision-aided motion estimation framework for the 
Roboray humanoid robot by integrating visual-kinematic odometry with inertial measurements. 
Furthermore, 
they employed the pose estimation to reconstruct a 3D voxel map of the environment utilizing depth data 
from a Time-of-Flight camera. However, they did not integrate the depth data for localization purposes. 

Works such as \cite{wagner2014graph} and 
\cite{scona2017direct} leveraged the frame-to-model visual tracking of 
KinectFusion~\cite{newcombe2011kinectfusion} or ElasticFusion~\cite{whelan2015elasticfusion}, which 
use a coarse-to-fine ICP algorithm. Nevertheless, vision-based SLAM techniques struggle in 
varying illumination conditions. While this has been explored in works such as~\cite{milford2012seqslam} and \cite{porav2018adversarial}, it remains an 
open area of research. In addition, the relatively short range of visual observations limit their 
performance for large-scale operations.

A probabilistic localization approach exploiting LiDAR information was implemented by Hornung~\textit{et al.} \cite{hornung2010humanoid} on the
Nao humanoid robot. Utilizing a Bayes filtering method, they integrated the measurements 
from a 2D range finder with a motion model to recursively estimate the pose of the robot's 
torso. They did not address the SLAM problem directly since they assumed the availability of a volumetric map for step planning.

Compared to bipedal robots, quadrupedal robots have better 
versatility and resilience when navigating challenging terrain. Thus, 
they are more suited to long-term tasks such as inspection of industrial sites. 
In the extreme case, a robot might need to jump over a terrain hurdle.
Park~\textit{et al.}, in \cite{park2015online}, used a Hokuyo laser range 
finder on the MIT Cheetah 2 to detect the ground plane as well as the front face of a hurdle. 
Although, the robot could jump over hurdles as high as
40 cm, the LiDAR measurements were not exploited to contribute 
to the robot's localization.  

Nobili~\textit{et al.} \cite{nobili2017heterogeneous} presented a state estimator for 
the Hydraulic Quadruped robot which took advantage of a variety of sensors, 
including inertial, kinematic and LiDAR measurements. Based on a modular inertial-driven EKF, 
the robot's base link velocity and position were propagated using measurements 
from modules including leg odometry, visual odometry and LiDAR odometry. Nevertheless, 
the system is likely to suffer from drift over the course of a large trajectory, as the
navigation system did not include a mechanism to detect loop-closures. 

\begin{figure}
\vspace{3mm}
\includegraphics[width=\linewidth,trim={0cm 2cm 27cm 17cm},clip]{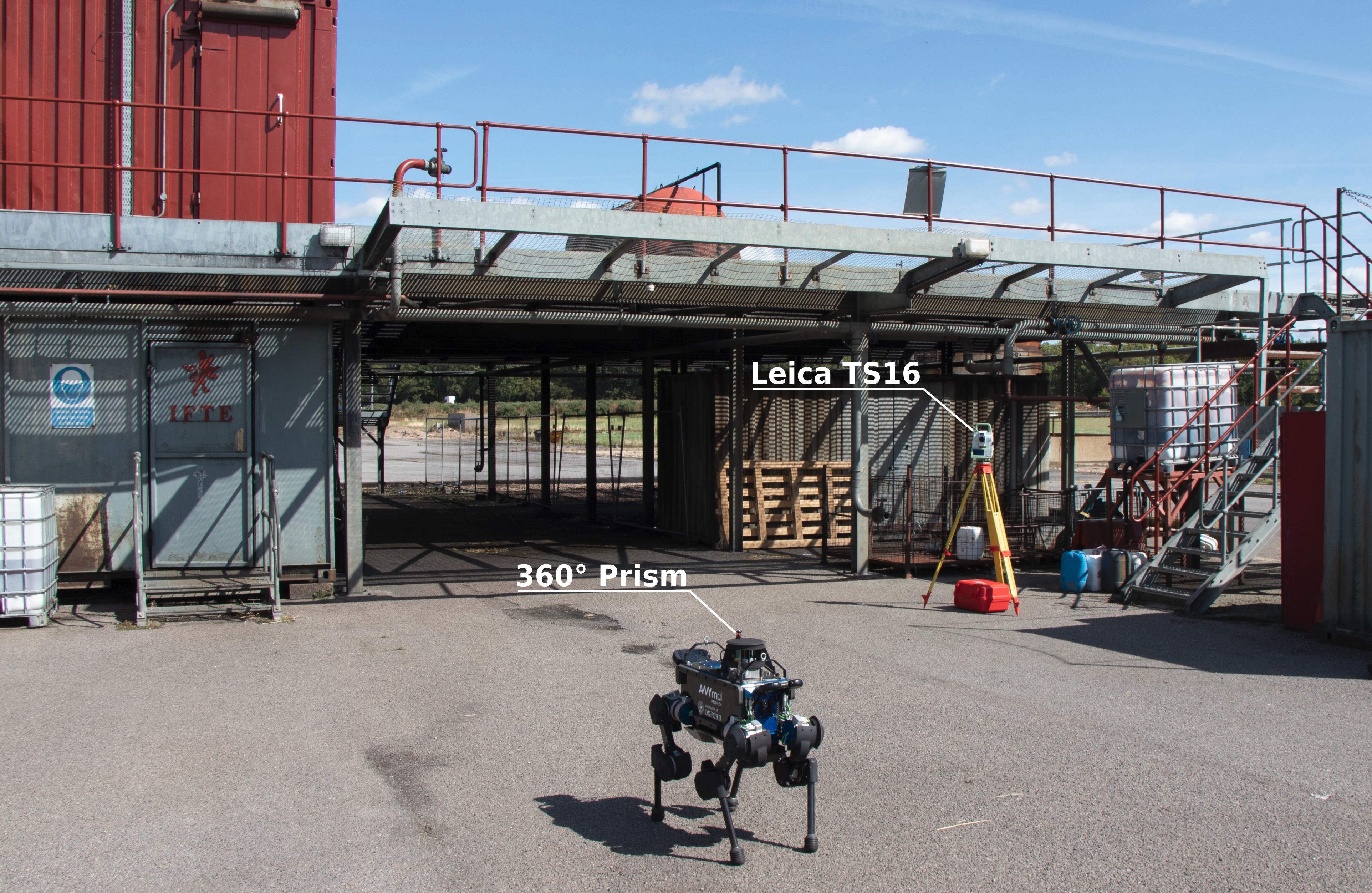}
\caption{\small{Our experiments were carried out with the quadrupedal robot, ANYmal, in the 
\textit{Oil Rig} training site at the Fire Service College, 
Moreton-in-Marsh, UK. To determine ground truth robot poses, we used a Leica TS16 laser 
tracker to track a $360^{\circ}$ prism on the robot.}}
\label{fig:Anymal}
\vspace{-2em}
\vspace{-3mm}
\end{figure}
%\vspace{-0.5em}
\subsection{LiDAR-SLAM Systems}
%\vspace{-0.5em}
The well-known method for point cloud registration is ICP which is used in a variety of 
application~\cite{pomerleau2013applied}. Given an initialization for the sensor pose, ICP iteratively 
estimates the relative transformation between two point clouds. A standard formulation of ICP minimizes 
the distance 
between corresponding points or planes after data filtering and outlier rejection. However, ICP suffers 
from providing a good solution in ill-conditioned situations such as when the robot crosses a door way.
 
An effective LiDAR-based approach is LiDAR Odometry and Mapping (LOAM)~\cite{zhang2014loam}. 
LOAM extracts edge and surface point features in a LiDAR cloud by 
evaluating the roughness of the local surface. The features are reprojected to the start of the next scan 
based on a motion model, with point correspondences found within 
this next scan. Finally, the 3D motion is estimated recursively by minimizing the overall 
distances between point correspondences. LOAM achieves high accuracy and low cost of computation.

Shan~\textit{et al.} in \cite{shan2018lego} extended LOAM with Lightweight and Ground-Optimized LiDAR 
Odometry and Mapping (LeGO-LOAM) which
added latitudinal and longitudinal parameters separately in a two-phase optimization. In addition, 
LeGO-LOAM 
detected loop-closures using an ICP algorithm to create a globally consistent pose graph using 
iSAM2~\cite{kaess2012isam2}.

Dub\'{e}~\textit{et al.} in \cite{dube2017online} developed an online cooperative LiDAR-SLAM system. Using two and three robots, each 
equipped with 3D LiDAR, an incremental sparse pose-graph is populated by successive place 
recognition constraints. The place recognition constraints are identified utilizing the SegMatch 
algorithm~\cite{dube2017segmatch}, which represents LiDAR clouds as a set of 
compact yet discriminative features. The descriptors were used in a matching process to
determine loop closures.

Our approach is most closely related to the localization system in \cite{nobili2017heterogeneous}. 
However, in order to have a globally consistent map and to maintain drift-free global localization, 
we build a LiDAR-SLAM on top of the AICP framework which enables detection of loop-closures and adds 
these constraints to a factor-graph.
Furthermore, we develop a fast 
verification of point-cloud registrations to effectively avoid incorrect factors to be added to the pose graph. 
We explain each component of our work in~\secref{sec:approach}.

\begin{figure}
\centering
\includegraphics[width=0.9\linewidth]{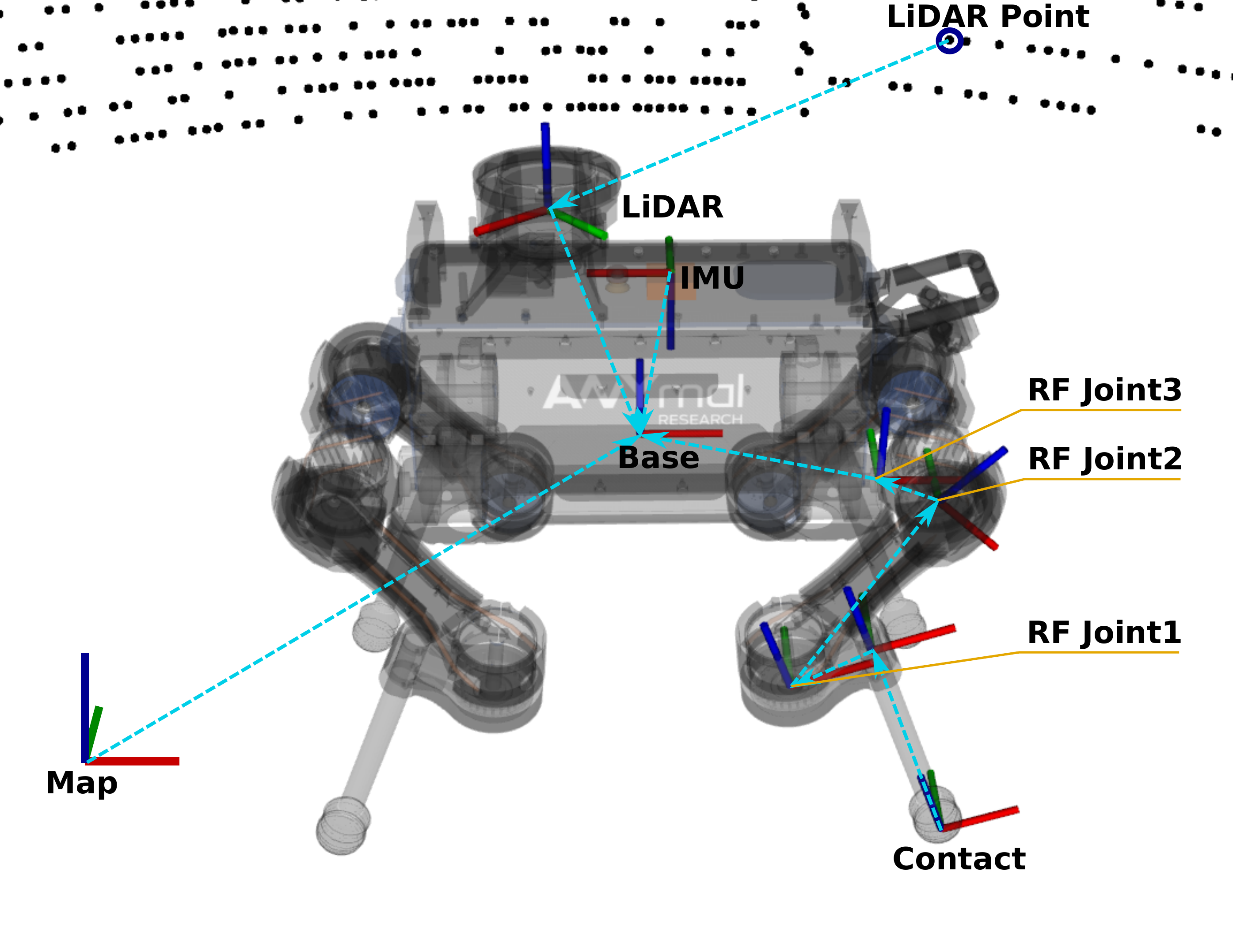}
\vspace{-1em}
\caption{\small{Axis conventions of various frames used in the 
LiDAR-SLAM system and their relationship with respect to the base frame. As an 
example, we show only the Right Front (RF) leg.}}
\label{fig:anymal_frames}
\vspace{-2em}
\end{figure}

%\vspace{-0.4em}
\section{Platform and Experimental Scenario} 
%\vspace{-0.5em}
\label{sec:scenario}
We employ a state-of-the-art quadruped, ANYmal (version B) \cite{hutter2016anymal}, as our experimental 
platform.~\figref{fig:Anymal} shows a view of the ANYmal robot. The robot weighs about $33$~\si{kg} without any external perception modules and can 
carry a maximum payload of $10$~\si{kg} at the maximum speed of $1.0$~\si{m/s}. As shown 
in~\figref{fig:anymal_frames}, each leg contains 3 actuated joints which altogether gives 
18 degrees of freedom, 12 actuated joints and the 6 DoF robot base, which the robot uses to 
dynamically navigate challenging terrain.

\tabref{table:Sensors} 
summarizes the specifications of the robot's sensors. The high 
frequency sensors (IMU, joint encoders and torque sensors) are tightly coupled through a 
Kalman filtering approach \cite{bloesch2017two} on the robot's Locomotion Personal 
Computer (LPC). Navigation measurements from the Velodyne VLP-16 LiDAR sensor and the robot's 
cameras are processed on a separate on-board computer to achieve online localization and mapping, as 
well as other tasks such as terrain mapping and obstacle avoidance. The pose estimate of our 
LiDAR-SLAM system is fed back to the LPC for path planning. 

\begin{table}[b]
	\centering
	\resizebox{\columnwidth}{!}{%
	\begin{tabular}{|c|cll|}
		\hline 
		\cellcolor{orange!55}\textbf{Sensor} & \cellcolor{orange!55}\textbf{Model} & \cellcolor{orange!55}\textbf{Frequency} & \cellcolor{orange!55}\textbf{Specifications} 
 
		\\ \hline \hline 		
		\multirow{4}{*}{\cellcolor{gray!35}}& & &\\ 
		 \cellcolor{gray!35}&  &  &Bias Repeatability:~$<0.5^\circ/s$; $~5$ $mg$\\
		 \cellcolor{gray!35}\multirow{-2}{*}{\textbf{IMU}}& \multirow{-3}{*}{Xsens}  & \multirow{-2}{*}{$400$} &\\
		 \cellcolor{gray!35}&\multirow{-3}{*}{MTi-100} & & \multirow{-2}{*}{Bias Stability: $10^\circ/h$;~$40$ $\mu g$} \\      
         \hline
        \cellcolor{gray!35} && &  \\ 
		\multirow{3}{*}{\cellcolor{gray!35}} & \multirow{3}{*}{Velodyne} & \multirow{4}{*}{$10$} &Resolution in Azimuth: $<0.4^\circ$	 \\ 
		\cellcolor{gray!35}\textbf{LiDAR} & \multirow{3}{*}{VLP-16} &  & Resolution in Zenith: $2.0^\circ$  	 \\ 
		\cellcolor{gray!35}\textbf{Unit}&  &  & Range~$<100$ $m$  \\
		\cellcolor{gray!35}&  &  & Accuracy: $\pm$ $3$ $cm$   \\
		\cellcolor{gray!35} && &  \\
		\hline
		\cellcolor{gray!35} && &  \\ 
		\cellcolor{gray!35}\multirow{-2}{*}{\textbf{Encoder}}&\multirow{-2}{*}{ANYdrive}&\multirow{-2}{*}{$400$}&\multirow{-2}{*}{Resolution~$<0.025^\circ$} \\
		\hline 
		\cellcolor{gray!35}  & &  &\\ 
		\cellcolor{gray!35}\multirow{-2}{*}{\textbf{Torque}}& \multirow{-2}{*}{ANYdrive}&\multirow{-2}{*}{$400$} &\multirow{-2}{*}{Resolution~$<0.1$ $N m$} \\ 
		\hline
	\end{tabular}
	}
	\caption{ \small{Specifications of the sensors installed on the ANYmal.}}
	\label{table:Sensors}
	\vspace{-3mm}
\end{table}  
%\vspace{-0.4em}
\section{Approach}
\begin{figure}[t]
\centering	
\resizebox{\columnwidth}{!}{%	
\includegraphics[width=8.5cm]{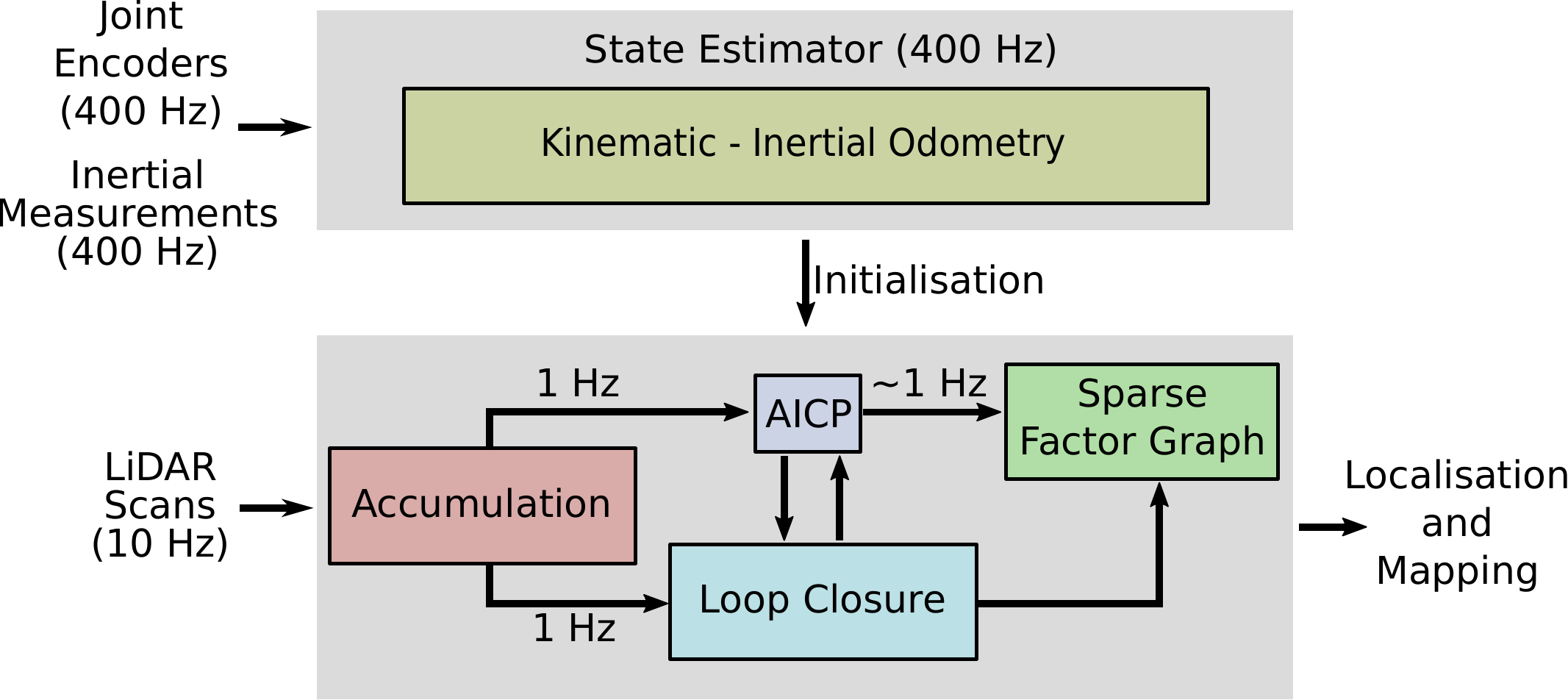}
}	
\caption{\small{Block diagram of the LiDAR-SLAM system.}}
\label{fig:diagram}
\vspace{-2em}
\end{figure}

%\vspace{-0.5em}
\label{sec:approach}
Our goal is to provide the quadrupedal robot with a drift-free localization estimate over the 
course of a very long mission, as well as to enable the robot to accurately map its surroundings. The 
on-the-fly map can further be employed for the purpose of optimal path planning as edges on the factor 
graph also indicate safe and traversable terrain.~\figref{fig:diagram} elucidates the different components 
of our system.

We correct the pose estimate of the kinematic-inertial odometry by the LiDAR data association in a 
loosely-coupled fashion. 

%\vspace{-0.4em}
\subsection{Kinematic-Inertial Odometry}
%\vspace{-0.5em}
The proprioceptive sensors are tightly coupled using a recursive error-state estimator, Two-State Implicit 
Filter (TSIF) \cite{bloesch2017two}, which estimates the incremental motion of the robot. By 
assuming that each leg is in stationary contact with the terrain, the position of the contact feet in 
the inertial frame $\{\mathcal{I}\}$, obtained from forward kinematics, is treated as a temporal 
measurement to estimate the robot's pose in the fixed odometry frame, 
$\{\mathcal{O}\}$:
\begin{equation}
\textbf{T}_{\mathcal{B}}^{\mathcal{O}} = \begin{bmatrix}
\textbf{R}_{\mathcal{B}}^{\mathcal{O}} & \textbf{p}_{\mathcal{B}}^{\mathcal{O}} \\
\text{0} & 1 
\end{bmatrix},
\end{equation}
where $\textbf{T}_{\mathcal{B}}^{\mathcal{O}} \in SE(3)$ is the transformation from the base frame 
$\{\mathcal{B}\}$ to the odometry frame $\{\mathcal{O}\}$.

This 
estimate of the base frame drifts over time as it does not use any exteroceptive sensors and the 
position of the quadruped and its rotation around $z$-axis are 
not observable. In the following, we define our LiDAR-SLAM system for estimation of the robot's pose with 
respect to the map frame $\{\mathcal{M}\}$ which is our goal. It is worth noting that the TSIF 
framework estimates the covariance of the state~\cite{bloesch2017two} which we employ during our geometric loop-closure detection. 
%\vspace{-0.4em}
\subsection{LiDAR-SLAM System}
%\vspace{-0.5em}
Our LiDAR-SLAM system is a pose graph SLAM system built upon our ICP registration approach called
Autotuned-ICP (AICP) \cite{Nobili17icra}. AICP automatically adjusts the outlier filter of 
ICP by computing an overlap parameter, $\Omega\in [0,1]$ since the 
assumption of a constant overlap, which is conventional in the standard outlier filters, violates the 
registration in real scenarios. 

Given the initial estimated pose from the kinematic-inertial odometry, we obtain a \textbf{
reference} cloud to which we align each consecutive \textbf{reading}\footnote{Borrowing the 
notation from\cite{pomerleau2013comparing}, the~\textbf{reference} and \textbf{reading} clouds 
correspond to the robot's poses at the start and end of each edge in the factor graph and AICP 
registers the latter to the former.} point cloud. 

In this manner the successive reading clouds are precisely aligned with the reference clouds with greatly
reduced drift. The robot's 
pose, corresponding to each reading cloud is obtained as follows:
\begin{equation}
\textbf{T}_{\mathcal{B}}^{\mathcal{M}} = \Delta_{aicp}\textbf{T}_{\mathcal{B}}^{\mathcal{O}},
\end{equation}     
where $\textbf{T}_{\mathcal{B}}^{\mathcal{M}}$ is the robot's pose in the map frame $\{\mathcal{M}\}$ 
and $\Delta_{aicp}$ is the alignment transformation calculated by AICP. 

Calculating the corrected poses, corresponding to the point clouds, we compute the relative 
transformation between the successive reference clouds to create the odometry factors of the 
factor graph which we introduce in~\eqref{eq:odom-factor}.

The odometry factor, $\phi_i(\mathcal{X}_{i-1},\mathcal{X}_{i})$, is defined as:
\begin{equation}
\label{eq:odom-factor}
\phi_{i}(\mathcal{X}_{i-1},\mathcal{X}_{i}) = (\textbf{T}_{\mathcal{B}_{i-1}}^{\mathcal{M}^{-1}}\textbf{T}_{\mathcal{B}_i}^{\mathcal{M}})^{-1}
\widetilde{\textbf{T}}_{\mathcal{B}_{i-1}}^{\mathcal{M}^{-1}}\widetilde{\textbf{T}}_{\mathcal{B}_i}^{\mathcal{M}},
\end{equation}
where $\widetilde{\textbf{T}}_{\mathcal{B}_{i-1}}^{\mathcal{M}}$ and 
$\widetilde{\textbf{T}}_{\mathcal{B}_i}^{\mathcal{M}}$ are the AICP estimated poses 
of the robot for the node $\mathcal{X}_{i-1}$ and $\mathcal{X}_{i}$, respectively, and 
$\textbf{T}_{\mathcal{B}_{i-1}}^{\mathcal{M}}$ and $\textbf{T}_{\mathcal{B}_i}^{\mathcal{M}}$ are the 
noise-free transformations.

A prior factor, 
$\phi_0(\mathcal{X}_0)$, which is taken from the pose estimate of the kinematic-inertial odometry, 
is initially added to the factor graph to set an origin and a heading for the robot within the map
frame $\{\mathcal{M}\}$.

To correct for odometric drift, loop-closure factors are
added to the factor graph once the robot revisits an area of the environment. We implemented two
approaches for proposing loop-closures: a) geometric proposal based on the distance between the current pose 
and poses already in the factor graph which is useful for smaller environments, and b) a learning approach for 
global loop-closure proposal, detailed in Sec~\ref{sec:esm}, which scales to large environments.
Each proposal provides an initial guess, which is refined with ICP.

Each individual loop closure becomes a factor and is added to the factor graph. The loop-closure factor, in this 
work, is a factor whose end is the current pose of the robot and whose start is one of the 
reference clouds, stored in the history of the robot's excursion. The nominated reference cloud 
must meet two criteria of nearest neighbourhood and sufficient overlap with the current reference 
cloud. 
An accepted loop-closure factor, 
$\phi_{j}(\mathcal{X}_{M},\mathcal{X}_{N})$, is defined as:
\begin{equation}
\phi_{j}({\mathcal{X}_{M},\mathcal{X}_{N}}) = 
(\textbf{T}_{\mathcal{B}_M}^{\mathcal{M}^{-1}}\textbf{T}_{\mathcal{B}_{N}}^{\mathcal{M}})^{-1}(\Delta_{j,aicp}\widetilde{\textbf{T}}_{\mathcal{B}_M}^{\mathcal{M}})^{-1}\widetilde{\textbf{T}}_{\mathcal{B}_{N}}^{\mathcal{M}},
\end{equation}   
where $\widetilde{\textbf{T}}_{\mathcal{B}_M}^{\mathcal{M}}$ and $\widetilde{\textbf{T}}_{\mathcal{B}_{N}}^{\mathcal{M}}$ are the robot's poses in the 
map frame $\{\mathcal{M}\}$ corrected by AICP with respect to the reference cloud $(M-1)$ and the reference cloud $(N-1)$, respectively. The $\Delta_{j,aicp}$ is the AICP correction between the current 
reference cloud $M$ and the nominated reference cloud $N$. 

Once all the factors, including odometry and loop-closure factors, have been added to the factor 
graph, we optimize the graph so that we find the Maximum A Posteriori (MAP) estimate for the 
robot poses corresponding to the reference clouds. To carry out this inference 
over the variables $\mathcal{X}_i$, where $i$ is the number of the robot's pose in the factor 
graph, the product of all the factors, must be maximized:
\begin{equation}
\label{eq:map_max}
\mathcal{X}^{MAP} = \ \argmax_{\mathcal{X}} \ \prod_i \phi_i(\mathcal{X}_{i-1},\mathcal{X}_{i}) \prod_j 
\phi_j(\mathcal{X}_{M},\mathcal{X}_{N}).
\end{equation}

Assuming that factors follow a Gaussian distribution and all measurements are only corrupted 
with white noise, i.e. noise with normal distribution and zero mean, the optimization problem 
in~\eqref{eq:map_max} is equivalent to minimizing a sum of nonlinear least squares:
\begin{equation}
\label{eq:map_min}
\begin{split}
\mathcal{X}^{MAP} = \ \argmin_{\mathcal{X}} \sum_{i} 
||y_i(\mathcal{X}_i,\mathcal{X}_{i-1})-m_i||^2_{\bm{\Sigma}_i} \\ + \sum_{j} 
||y_j(\mathcal{X}_M,\mathcal{X}_{N})-m_j||^2_{\bm{\Sigma}_j},
\end{split}
\end{equation}
where $m$, $y$ and $\bm{\Sigma}$ denote the measurements, their mathematical model and the 
covariance matrices, respectively. 

As noted, the MAP estimate is only reliable when the residuals in~\eqref{eq:map_min} follow the normal distribution. However, ICP is susceptible to failure in 
the absence of geometric features, e.g. in corridors or door entries, which can have a
detrimental effect when optimizing the pose graph. In~\secref{sec:verification}, we propose a fast 
verification 
technique for point cloud registration to detect possible failure of the AICP registration. 
\setlength{\textfloatsep}{0pt}% Remove \textfloatsep
\begin{algorithm}[t]
\caption{ \small{Improved Risk Alignment Prediction.}}
\label{alg:verification}
\small{
\SetAlgoLined
\DontPrintSemicolon
\textbf{input}: source, target clouds $\mathcal{C}_S$, $\mathcal{C}_T$; estimated poses 
$\mathcal{X}_{S}$, $\mathcal{X}_{T}$\;
\textbf{output}: alignment risk $\rho = f(\Omega, \alpha)$\;
\Begin{
Segment $\mathcal{C}_S$ and $\mathcal{C}_T$ into a set of planes: $P_{Si}$ and $P_{Tj}$ $i \in N_S$, $j \in N_T$,\;
Compute centroid of each plane (Keypoint): $\bm{K}_{Si}$, $\bm{K}_{Tj}$,\;
Transform query keypoints $\bm{K}_{Tj}$ into the space of $\mathcal{C}_S$,\;
Search the nearest neighbour of each query plane $P_{Tj}$ using a k-d tree,\; 
\For{plane $P_{T}$}
{Find match $P_{S}$ amongst candidates in the k-d tree,\;
Compute the matching score $\Omega_p$,\;
\If{$\Omega_p$ is max}{Determine the normal of plane $P_T$,\;
Push back the normal into the matrix $\bm{N}$,}}
Compute alignability $\alpha = \lambda_{max}\mathbin{/}\lambda_{min}$;\;
Learn $\rho = f(\Omega, \alpha)$;\;
Return $\rho$;
}}
\end{algorithm}
%\vspace{-1.2em}
\subsection{Loop Proposal Methods}
%\vspace{-0.1em}
This section introduces our learned loop-closure proposal and geometric loop-closure detection.
\subsubsection{Deeply-learned Loop Closure Proposal}
\label{sec:esm}
We use the method of Tinchev~\textit{et al.}~\cite{tinchev2019learning}, which is based on matching
individual segments in pairs of point clouds using a deeply-learned feature descriptor. Its specific design 
uses a shallow network such that it does not 
require a GPU during run-time inference on the robot. We present a summary of the method called 
Efficient Segment Matching (ESM), but refer the reader to~\cite{tinchev2019learning}.

First, a neural network is trained offline using individual LiDAR observations (segments). By leveraging 
odometry in the process, we can
match segment instances without manual intervention. The input to the network 
is a batch of triplets - anchor, positive and negative segments. The anchor and positive 
samples are the same object from two successive Velodyne scans, while the negative segment is a 
segment chosen $\approx 20\,\text{m}$ apart. The method then performs a series of $X\text{-conv}$ 
operators directly on raw point cloud data, based on PointCNN~\cite{li2018pointcnn}, followed by three 
fully connected layers, where the last layer is used as the descriptor for the 
segments.

During our trials, when the SLAM system receives a new reference cloud, it is preprocessed and then
segmented into a collection of point cloud clusters. These clusters are not semantically 
meaningful, but they broadly correspond to physical objects such as a vehicle, a tree or building facade.
For each segment in the reference cloud, a descriptor 
vector is computed with an efficient TensorFlow C++ implementation by performing a forward pass using 
the weights from the already trained model. This allows a batch of 
segments to be preprocessed simultaneously with zero-meaning and normalized variance and then 
forward passed through the trained model. We 
use a three dimensional tensor as input to the network - the length is the number of segments in 
the current point cloud, the width represents a fixed-length down-sampled vector of all the 
points in an individual segment, and the height contains the $x$, $y$ and $z$ values. Due to the 
efficiency of the method, we need not split the tensor into
mini-batches, allowing us to process the full reference cloud in a single forward pass.

Once the descriptors for the reference cloud are computed, they are compared to the 
map of previous reference clouds. ESM uses an $l_2$ distance 
in feature space to detect matching segments and a 
robust estimator to retrieve a 6DoF pose. This produces a transformation of the current reference 
cloud with respect to the previous reference clouds. The transformation is then used in AICP to add a 
loop-closure as a constraint to the graph-based optimization. Finally, ESM's map representation is 
updated, when the optimization concludes.

\subsubsection{Geometric Loop-Closure Detection}
To geometrically detect loop-closures, we use the covariance of the legged state estimator (TSIF) to define a dynamic 
search window around the current pose of the robot. Then the previous robot's poses, which reside 
within the search window, are examined based on two criteria: nearest neighbourhood and verification of 
cloud registration (described in~\secref{sec:verification}). Finally, the geometric loop closure is computed 
between the current cloud and the cloud corresponding to the nominated pose using AICP. 

\begin{figure*}[t]
\centering
\begin{multicols}{2}
\includegraphics[width=8.7cm,height=4.3cm,trim={1.3cm 0cm 1.5cm 0.5cm},clip]{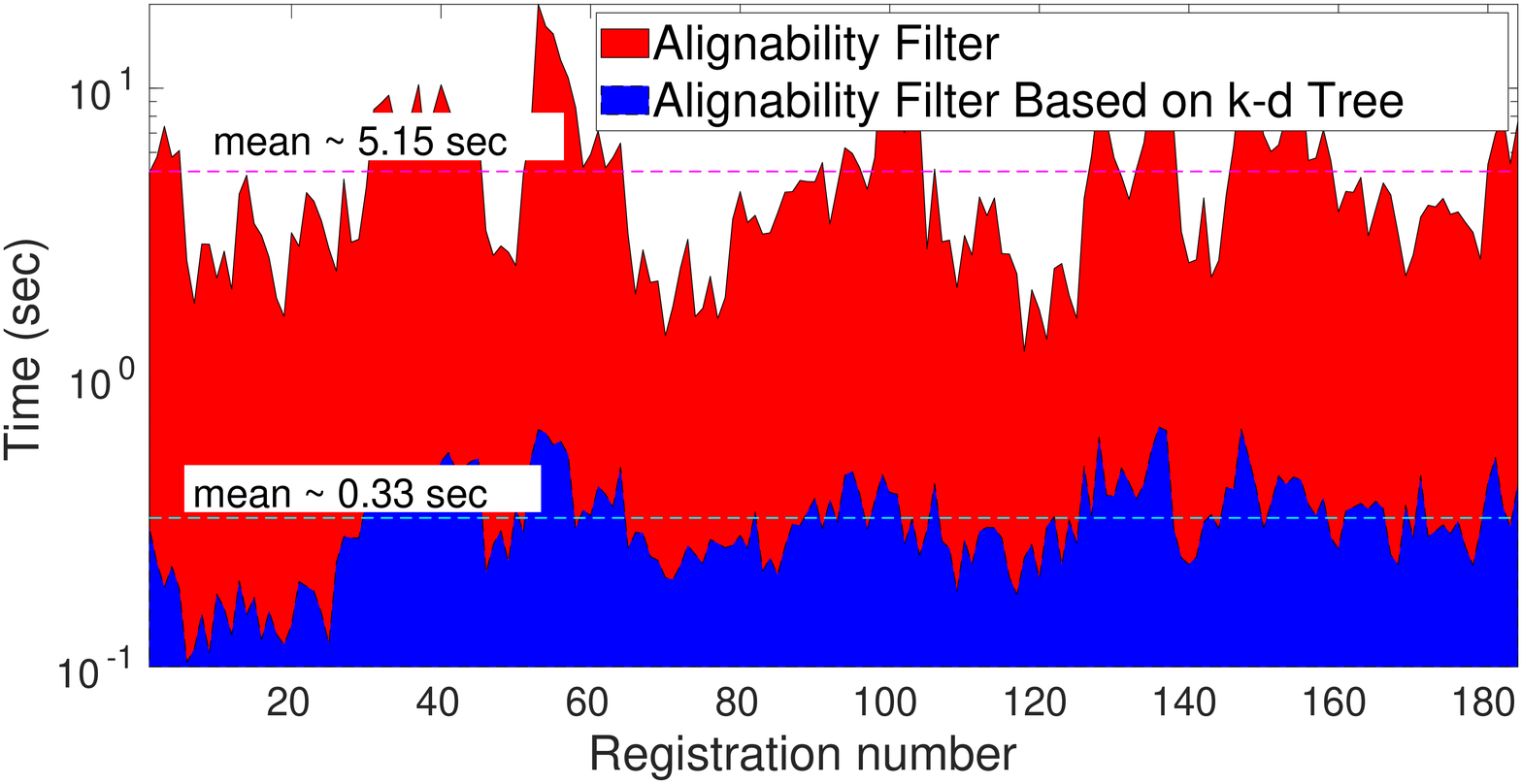}%
\includegraphics[width=8.7cm,height=4.3cm,trim={1.3cm 0cm 1.5cm 0.5cm},clip]{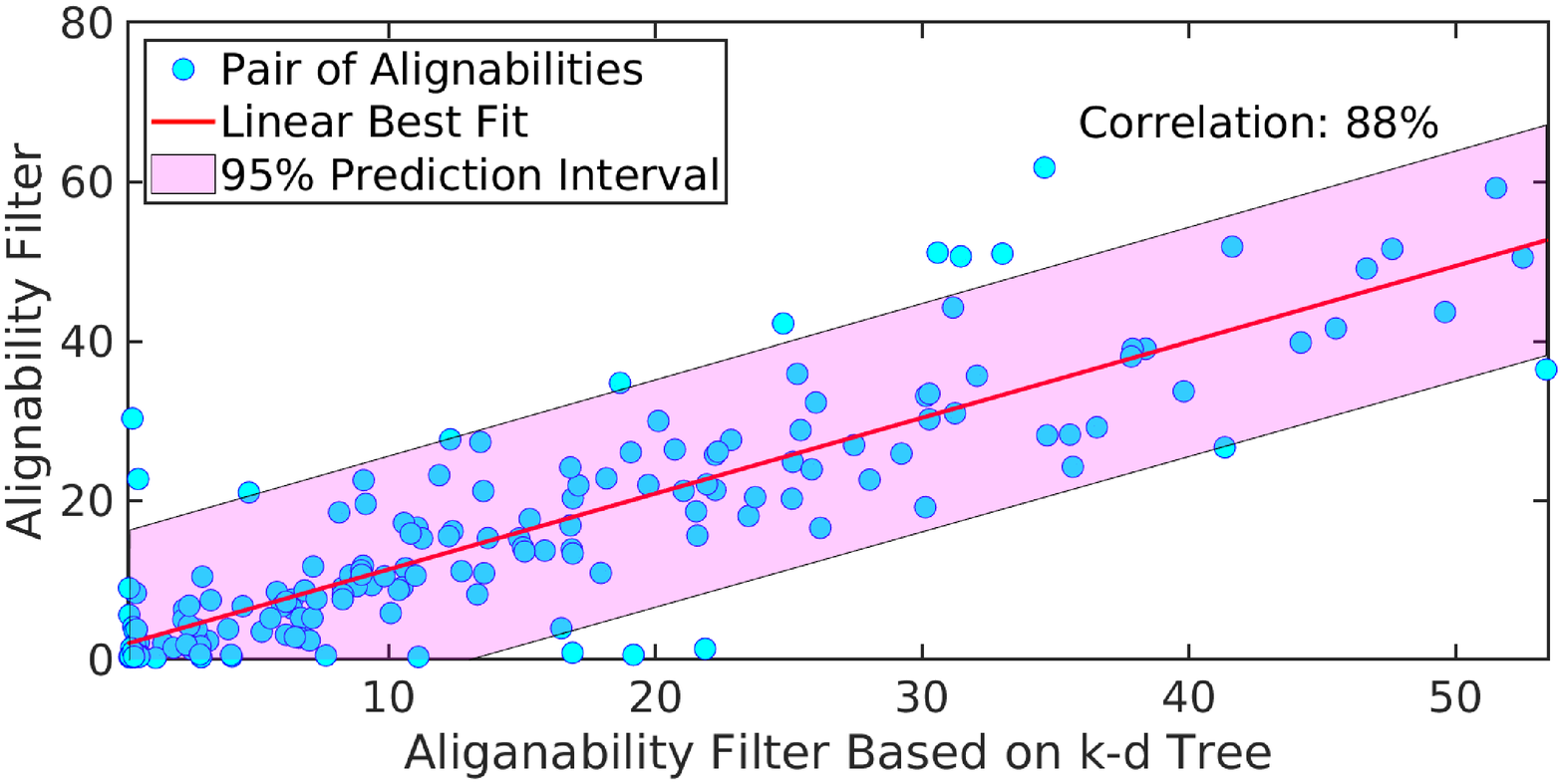}
\end{multicols}
\vspace{-1em}
\caption{\small{Comparison of our proposed verification method with the original in terms of computation time (left) and 
performance (right).}}
\label{fig:verification}
%\vspace{-2em}
\end{figure*}

\subsection{Fast Verification of Point Cloud Registration}
%\vspace{-0.3em}
\label{sec:verification}
This section details a verification approach for ICP point cloud registration to determine
if two point clouds can be safely registered. We improve upon our previously proposed alignability metric 
in~\cite{Nobili18icra} with a much faster method.

\textit{Method from} \cite{Nobili18icra}: First, the point clouds are segmented into a set of 
planar surfaces. Second, a matrix $\bm{N} \in \mathbb{R}
^{M\times 3}$ is computed, where each row corresponds to the normal of the planes ordered by
overlap. $M$ is the number of matching planes in the overlap region between the two 
clouds. Finally, the alignability metric, $\alpha$ is defined as the ratio between the smallest and 
largest eigenvalues of $\bm{N}$. 

The matching score, $\Omega_p$, is computed as the overlap between two planes, $P_{Tj}$ and $P_{Si}
$ where $i \in N_S$ and $j \in N_T$. $N_S$ and $N_T$ are the number of planes in the input clouds. 
In addition, in order to find the highest possible overlap, the algorithm iterates over all possible 
planes from two point clouds. This results in overall complexity $O(N_SN_T(N_{P_S}N_{P_T}))$, 
where $N_{P_S}$ and $N_{P_T}$ are the average number of points in planes of the two clouds.

\textit{Proposed Improvement:} To reduce the pointwise computation, we first compute the centroids of each plane $\bm{K}_{Si}$ 
and $\bm{K}_{Tj}$ and align them from point cloud $\mathcal{C}_T$ to $\mathcal{C}_S$ given the 
computed transformation. We then store the centroids in a k-d tree and for each query plane $P_T 
\in \mathcal{C}_T$ we find the $K$ nearest neighbours. We compute the overlap for the $K$ nearest 
neighbours, and use the one with the highest overlap. This results in $O(N_TK(N_{P_S}N_{P_T}))$, 
where $K << N_s$. In practice, we found that $K=1$ is sufficient for our experiments. Furthermore, we 
only store the centroids in a k-d tree, reducing the space complexity.

We discuss the performance of this algorithm, as well as its computational complexity in the 
experiment section of this paper. Pseudo code of the algorithm is available in 
Algorithm~\ref{alg:verification}.

\begin{figure}[b!]
	\includegraphics[width=\linewidth]{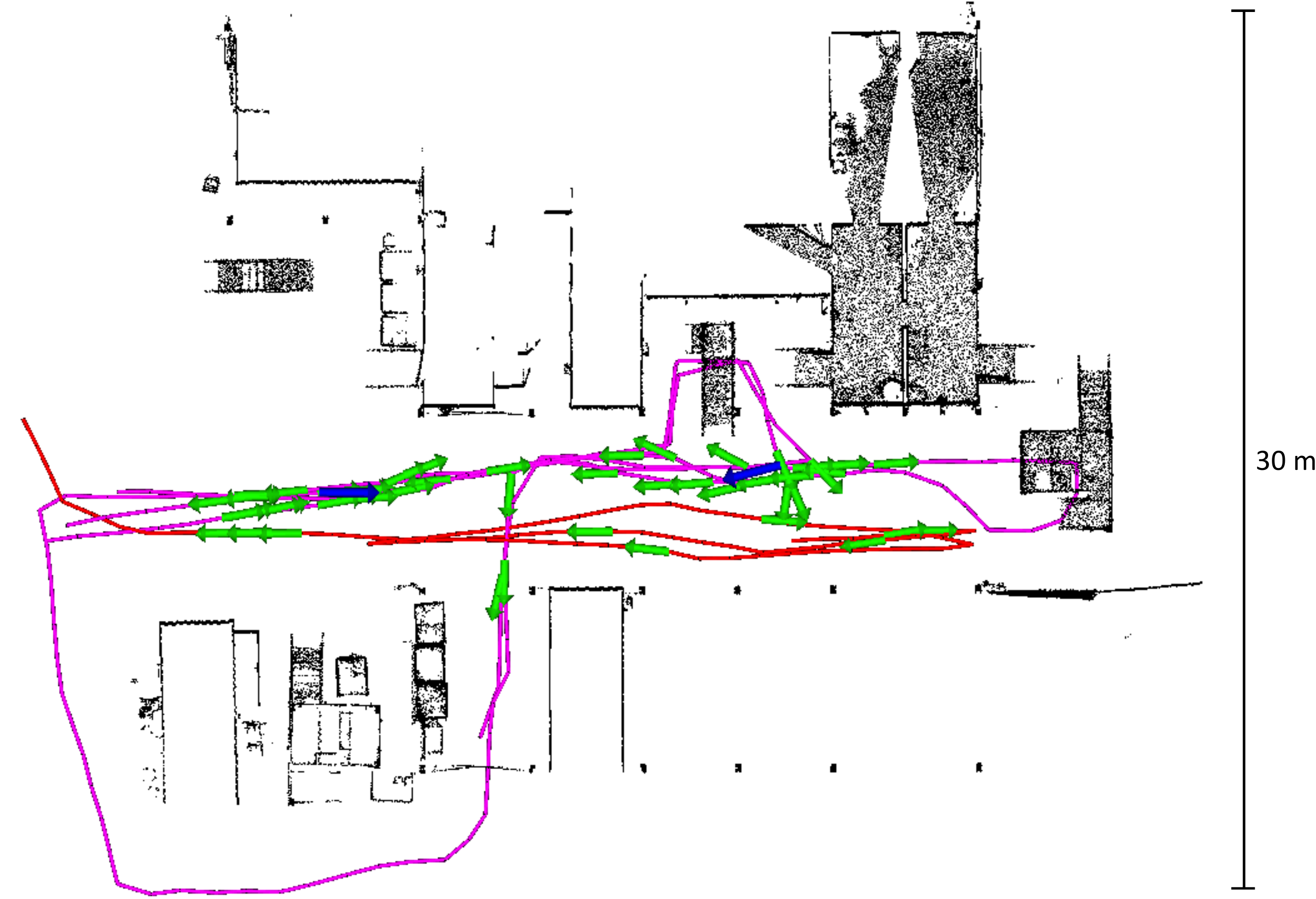}
	\caption{\small{Demonstration of the learning-based loop closure in the outdoor experiment. 249 point 
	clouds from two different runs were first registered against a ground truth Leica map (black). 
	The red and magenta traces correspond to the traversed path of each run. 
	Just two Velodyne point clouds are used to form a map for ESM (blue arrows). 
	Successful loop closures from the 50 clouds (green arrows) demonstrate
	robustness to viewpoint variation/offset.}}
	\label{fig:esm}
	\vspace{-1.5em}
\end{figure}
%\vspace{-0.4em}
\section{Experimental Evaluation}
%\vspace{-0.5em}
\label{sec:exp-eval}
The proposed LiDAR-SLAM system is evaluated using the datasets collected by our ANYmal quadruped robot. We first 
analyze the verification method. Second, we investigate the 
learned loop-closure detection in terms of speed and reliability (\secref{sec:esm_eval}). We then demonstrate 
the performance or our SLAM system on two large-scale experiments, one indoor and one outdoor (\secref{sec:exps})
including an online demonstration where the map is used for route following (\secref{sec:expOnRobot}). A demonstration 
video can be found at\\ \url{https://ori.ox.ac.uk/lidar-slam}.
%\vspace{-0.4em}
\subsection{Verification Performance}
%\vspace{-0.5em}
\label{sec:verification_results}
We focus on the alignability metric $\alpha$ of the alignment risk prediction since it is our primary 
contribution. We computed $\alpha$ between consecutive point clouds of our outdoor experiment, which 
we discuss later in~\secref{sec:exps}. As seen in~\figref{fig:verification} (Left), the alignability filter 
based on a k-d tree is substantially faster than the original filter. As noted in~\secref{sec:verification}, 
our approach is less dependent on the point cloud size, due to only using the plane centroids. Whereas, 
the original alignability filter fully depends on all the points of the segments, 
resulting in higher computation time. Having tested our alignability filter for the datasets taken from Velodyne 
VLP-16, the average computation time is less than 0.5 seconds (almost 15x improvement) which is 
suitable for real-time operation. 

\figref{fig:verification} (Right) shows that our approach highly correlates with the result from the 
original approach. Finally, we refer the reader to the original work that provided a thorough 
comparison of alignment risk against Inverse Condition Number (ICN)~\cite{cheney2012numerical} 
and Degeneracy parameter~\cite{zhang2016degeneracy} in ill-conditioned scenarios. The former 
mathematically assessed the condition of the optimization, whereas the latter determines the degenerate 
dimension of the optimization.

%\begin{table}
%	\centering
%	\resizebox{\columnwidth}{!}{%
%		\begin{tabular}{|l|c|c|c|c|}
%			\hline 
%			\cellcolor{orange!75} & \cellcolor{orange!75}\textbf{Num of loop closures} & 
%			\cellcolor{orange!75}\textbf{Avg. num of segments} 
%			&\cellcolor{orange!75}\textbf{RMSE (m)} & 
%			\cellcolor{orange!75}\textbf{Avg. computation time [ms]} \\
%			\hline \hline 
%			\textbf{ESM} & 50 &  5.24 & $0.08\pm 0.02$ & 423.404 ms \\
%			\hline 
%		\end{tabular}
%	}
%	\caption{ \small{Results from running the deep learned loop closure proposal method. The 
%	computation time is measured as an average per point cloud.}}
%	\label{table:esm_results}
%\end{table}

\begin{figure*}[b]
\centering
\begin{multicols}{2}
    \includegraphics[width=8.5cm,trim={0cm 1cm 0cm 5.5cm},clip]{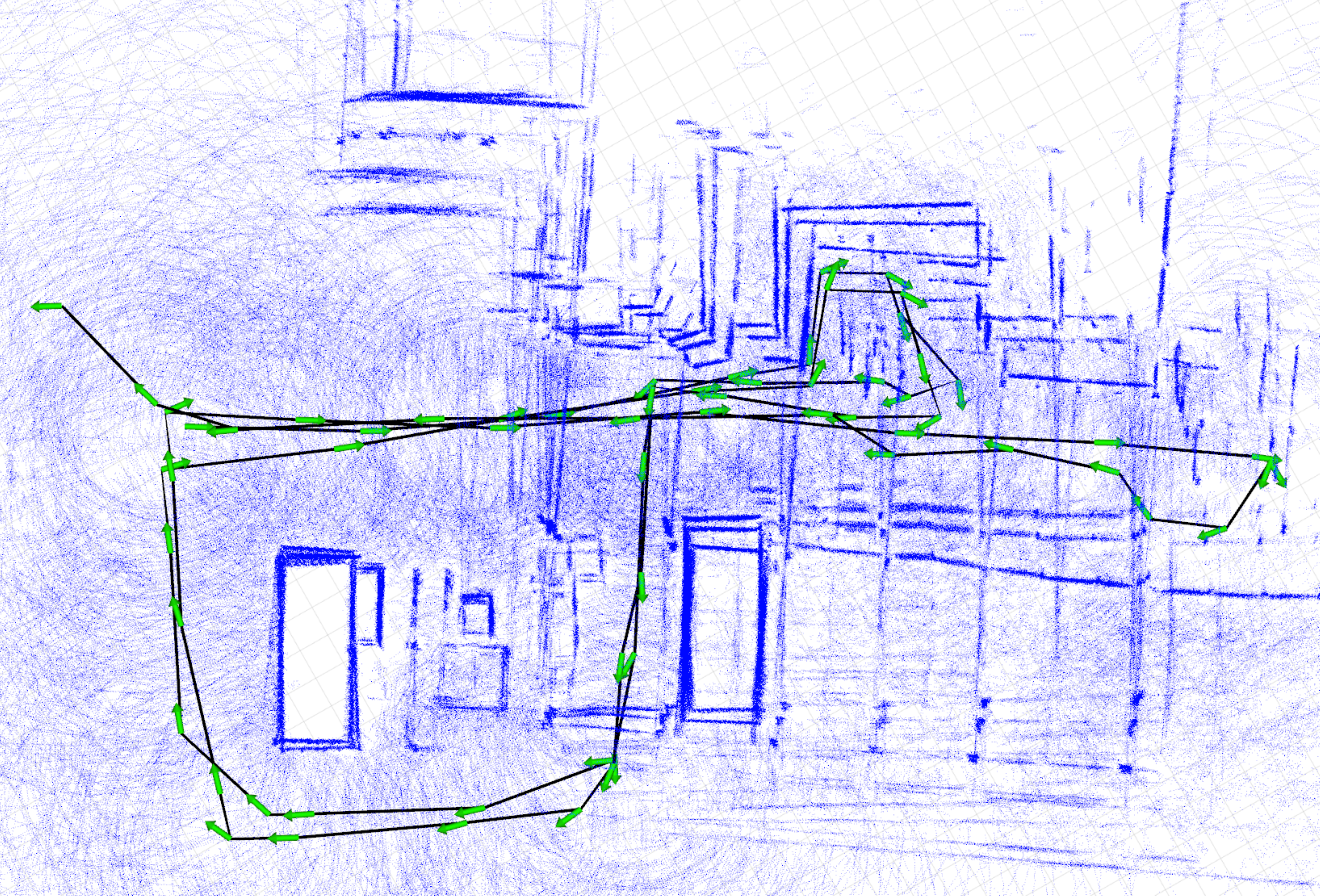}
     \centering \small{(a) AICP Odometry without loop-closure} 
    \includegraphics[width=8.5cm,trim={0cm 0cm 0cm 5.5cm},clip]{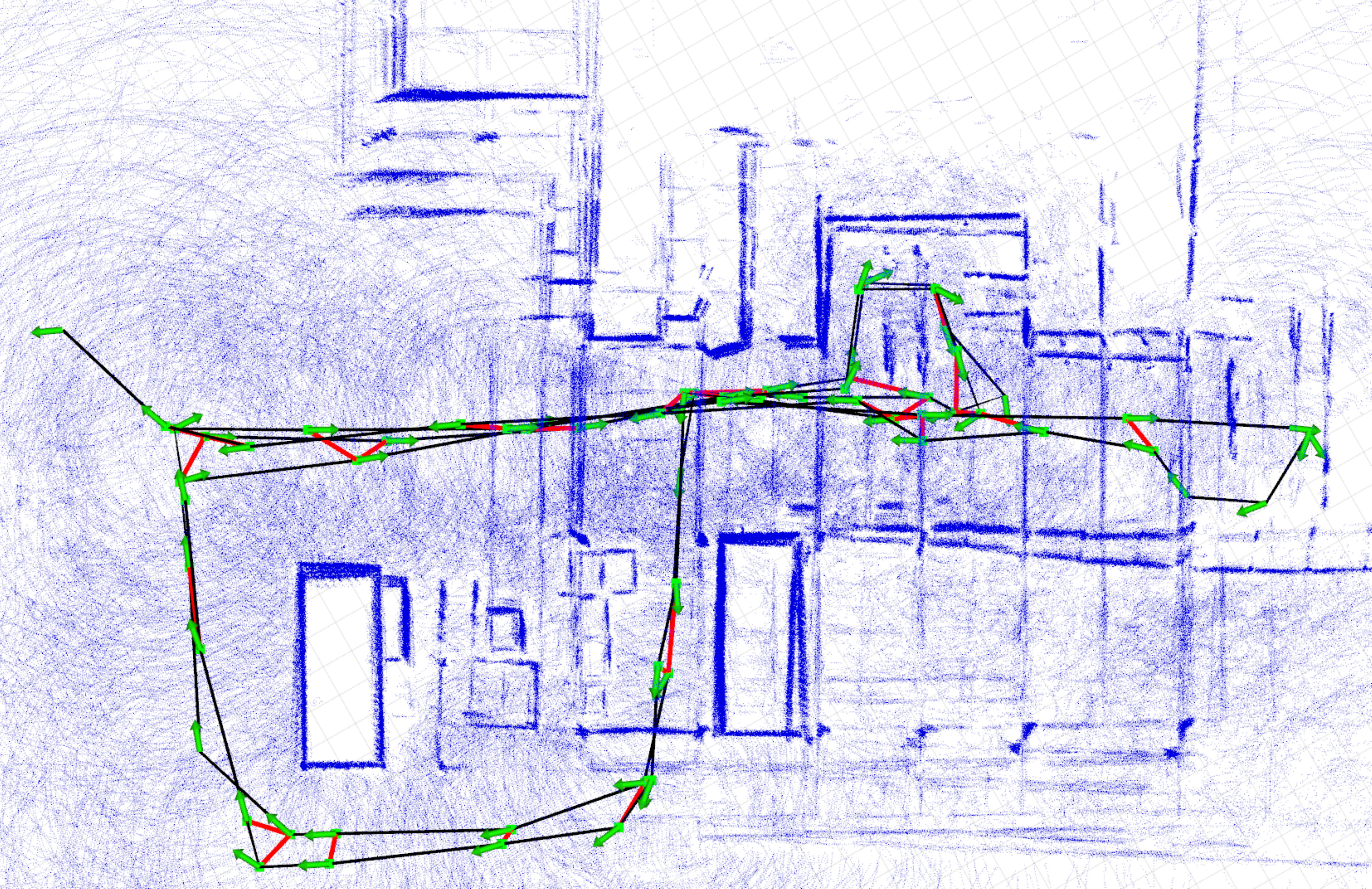}
     \centering \small{(b) SLAM without loop-closure verification}
\end{multicols}
\begin{multicols}{2}
    \includegraphics[width=8.5cm,trim={0cm 0cm 0cm 5cm},clip]{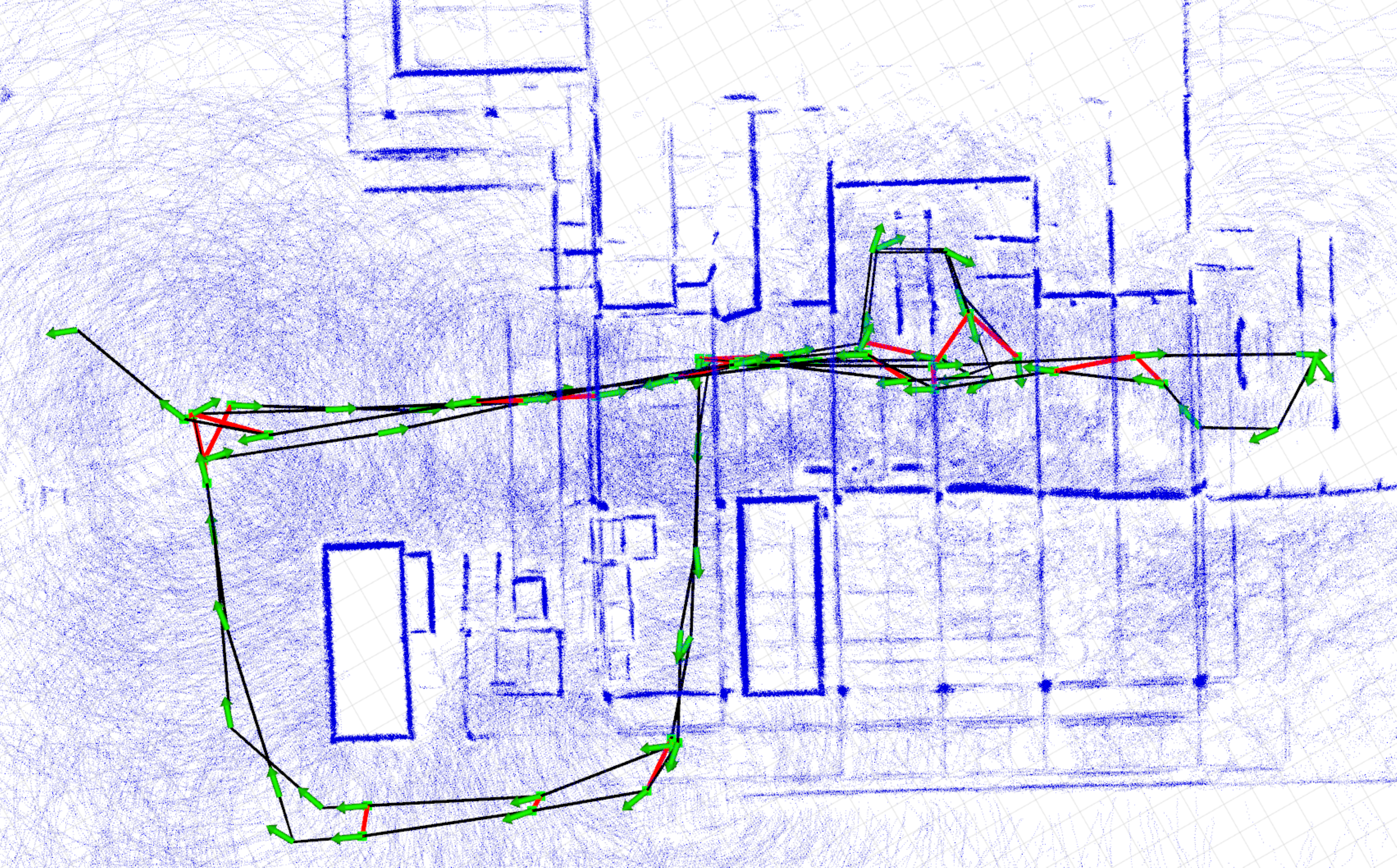}
     \centering \small{(c) SLAM with loop-closure verification}
    \includegraphics[width=8.5cm,trim={12.5cm 0cm 12.5cm 1.5cm},clip]{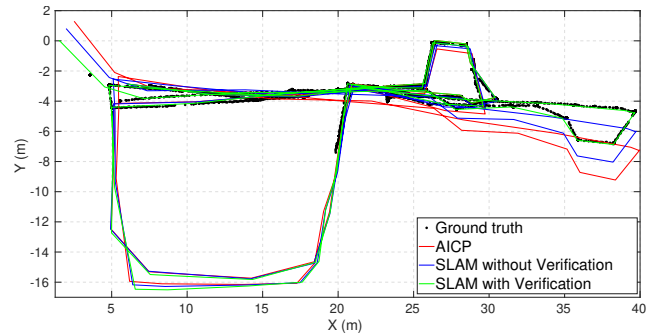}
     \centering \small{(d) Comparison against ground truth}
\end{multicols}
\vspace{-1em}
\caption{\small{Illustration of different algorithm variants: (a) AICP odometry, (b) the SLAM system without 
verification enabled (c) 
the SLAM system with it enabled, and finally (d) a comparison of their estimated trajectories with poses from Leica tracker (ground 
truth).}}
\label{fig:outdoor}
\vspace{-0em}
\end{figure*}

%\vspace{-0.4em}
\subsection{Evaluation of Learned Loop Closure Detector}
%\vspace{-0.5em}
\label{sec:esm_eval}

In the next experiment we explore the performance of the different loop closure methods
using a dataset collected outdoors at the Fire Service College, Moreton-in-Marsh, UK.

\subsubsection{Robustness to viewpoint variation}
\figref{fig:esm} shows a preview of our SLAM system with the learned loop closure method. We selected 
just two point clouds to create the map, with their positions indicated by the blue 
arrows in~\figref{fig:esm}.

The robot executed two runs in the environment, which comprised 249 point clouds. We deliberately 
chose to traverse an offset path (red) the second time so as to determine how robust our algorithm is to 
translation and viewpoint variation. In total 50 loop closures were detected (green arrows) around the 
two map point clouds. Interestingly, the approach not only detected loop closures from both 
trajectories, with translational offsets up to 6.5 m, but also with orientation variation up to $180^{\circ}$- 
something not 
achievable by standard visual localization - a primary motivation for using LiDAR. Across the 50 loop closures
an average of $5.24$ segments were recognised per point cloud. The 
computed transformation had an Root Mean Square Error (RMSE) of $0.08 \pm 0.02\,\text{m}$ from the 
ground truth alignment. 
This was achieved in approximately $486\,\text{ms}$ per query point cloud.
%
%\todo{this would be stronger if you could say that loop closures were detected with translational offsets 
%up to XXXm}
\begin{figure}[t]
\centering
	\includegraphics[width=\linewidth,trim={4.5cm 0cm 6cm 0.5cm},clip]{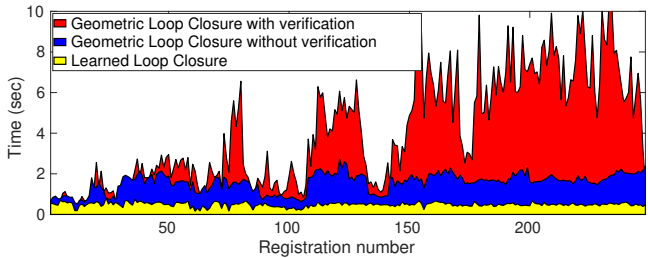}
	\caption{\small{Computation times of the considered loop closure methods. The computation time increases when the robot revisits old parts of the map, affecting the speed of geometric loop-closure detection, specifically when verification is enabled.}}
	\label{fig:time}
	\vspace{0em}
\end{figure}

\subsubsection{Computation Time}
~\figref{fig:time} shows a graph of the computation time. The computation time for the geometric loop closure method depends on the number of traversals around the same area. The geometric loop closures iterated over the nearest $N$ clouds, based on a radius; the covariance and distance travelled caused it to slow down. Similarly, the verification method needs to iterate over large proportion of the point clouds in the same area, affecting the real time operation.

Instead, the learning loop closure proposal scales better with map size. It compares low 
dimensional feature descriptor vectors, which is much faster than the thousands of data points in Euclidean space. 

%\vspace{-0.4em}
\subsection{Indoor and Outdoor Experiments}
%\vspace{-0.4em}
\label{sec:exps}
To evaluate the complete SLAM system, the robot walked indoor and outdoor along trajectories with the length of 
about 100 m and 250 m, respectively. Each experiment lasted about 45 minutes. \figref{fig:teaser} 
(Bottom) and \figref{fig:Anymal} illustrate the test locations: industrial buildings. For quality evaluation, we 
compared the SLAM system, AICP, and the legged odometry (TSIF) using ground truth. As shown 
in~\figref{fig:Anymal}, we used a Leica TS16 to automatically track a $360^{\circ}$ prism rigidly 
mounted on top of the robot. This way, we managed to record the robot's position with millimeter 
accuracy at about 7 Hz (when in line of sight). 

For evaluation 
metrics, we use Relative Pose Error (RPE) and Absolute Trajectory Error (ATE)~\cite{sturm2012benchmark}. 
RPE determines the regional accuracy of the trajectory over time. In other words, it is a metric which 
measures the drift of the estimated trajectory. ATE is the RMSE of 
the Euclidean distance between the estimated trajectory and the ground truth. ATE validates the 
global consistency of the estimated trajectory. 

As seen in~\figref{fig:outdoor}, our SLAM system is almost complete consistent with
the ground truth. The verification algorithm approved 27 
loop-closure factors (indicated in red) which were added to the factor graph. Without this verification
38 loop-closures were created, some in error, resulting in an inferior map.

\tabref{table:AteRunTimeStats} 
reports SLAM results with and without verification compared to the AICP LiDAR odometry and the TSIF legged state estimator. As the 
Leica TS16 does not provide rotational estimates, we took the best performing method - SLAM with 
verification - and compared the rest of the trajectories to it with the ATE metric. Based on this 
experiment, the drift of SLAM with verification is less than 0.07$\%$, satisfying many 
location-based tasks of the robot.

\begin{figure}[t]
\centering	
\resizebox{\columnwidth}{!}{%	
\includegraphics[width=8.5cm,height=6cm,trim={0cm 0cm 0cm 0cm},clip]{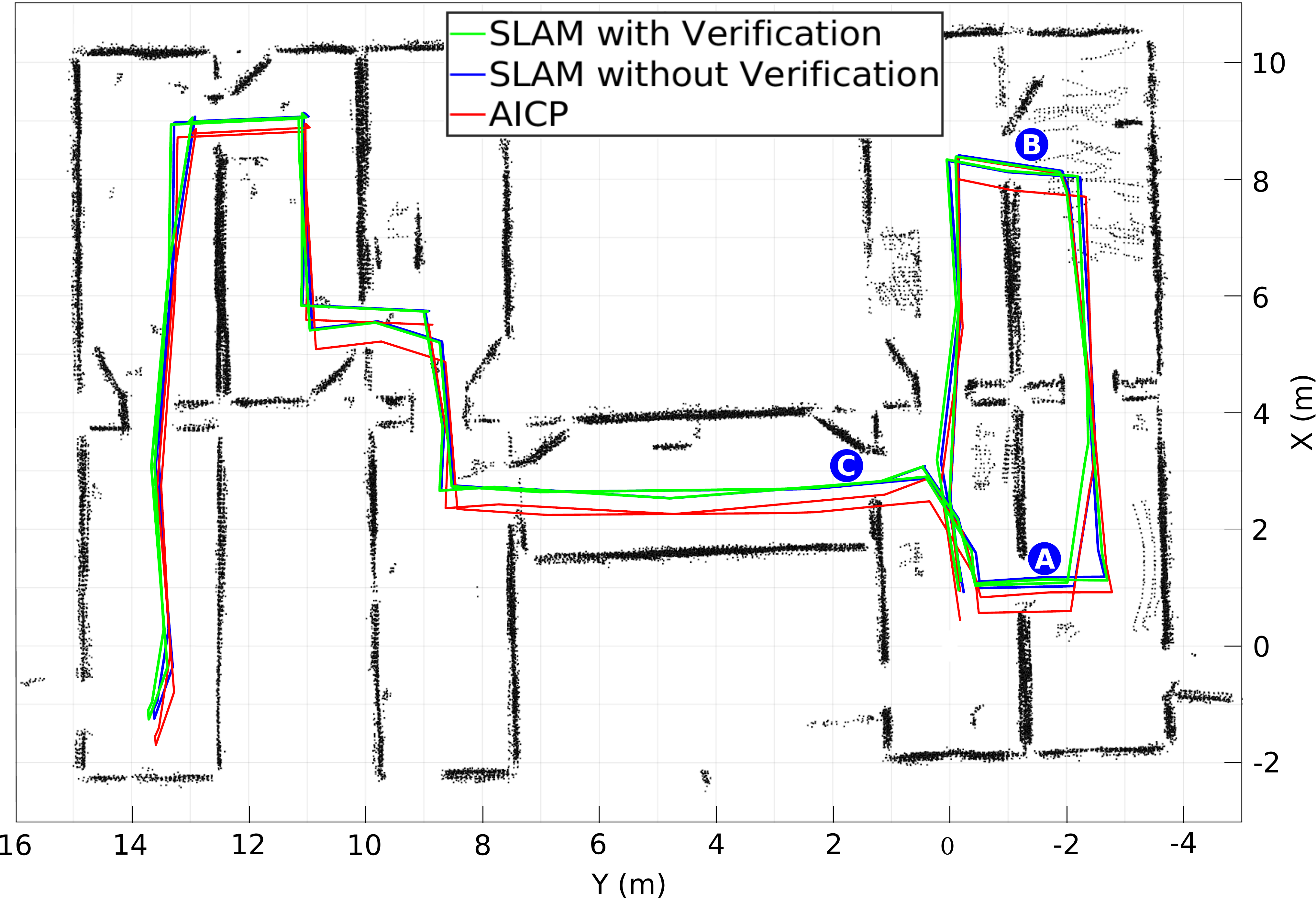}
}
\caption{\small{Estimated trajectories for the indoor experiment, overlayed on the map created by the SLAM system using the verification approach.}}
\label{fig:indoor}
%\vspace{-1.5em}
\end{figure}

For the indoor experiment, the robot walked along narrow passages with poor illumination and indistinct
structure. Due to the lack of ground truth, only estimated trajectories are only displayed 
in~\figref{fig:indoor}, with the map generated by SLAM with verification. As indicated, in points A, B 
and C, AICP approach cannot provide sufficient accuracy for the robot to safely pass through doorways 
about 1~m wide.

\begin{table}
	\centering
	\resizebox{\columnwidth}{!}{%
	\begin{tabular}{|l|c|c|c|}
		\hline 
		 \cellcolor{orange!55} & \cellcolor{orange!55}\textbf{Translational} &\cellcolor{orange!55}\textbf{Heading} & \cellcolor{orange!55}\textbf{Relative Pose} \\
		\multirow{-2}{*}{\cellcolor{orange!55}\textbf{Method}} &\cellcolor{orange!55}\textbf{Error (RMSE) (m)} & \cellcolor{orange!55}\textbf{Error (RMSE) (deg)} &\cellcolor{orange!55} \textbf{Error (m)} \\
        \hline \hline 
		\textbf{SLAM with Verification} & \cellcolor{green!75}\textbf{0.06}  & N/A &\cellcolor{green!75}\textbf{0.090} \\
		\hline 
		\textbf{SLAM without Verification} & \cellcolor{green!45}0.23  &\cellcolor{green!45}1.6840 
		&\cellcolor{green!45}0.640\\ 
		\hline 
		\textbf{AICP} (LiDAR odometry) & 0.62 &3.1950 &1.310  \\ 
		\hline 
        \textbf{TSIF} (legged odometry) & 5.40 &36.799 &13.64 \\ 
		\hline
	\end{tabular}
	}
	\caption{ \small{Comparison of the localization accuracy for the different approaches.}}
	\label{table:AteRunTimeStats}
%	\vspace{-2em}
\end{table}

%\vspace{-0.4em}
\subsection{Experiments on the ANYmal}
%\vspace{-0.5em}
\label{sec:expOnRobot}
In a final experiment, we tested the SLAM system online on the ANYmal. After building a map with several
loops (while teleoperated), we queried a path back to the operator station.
Using the Dijkstra's algorithm~\cite{dijkstra1959note}, the shortest path was created using the factor graph.
As each edge has previously been traverse, following the return trajectory returned the robot to the starting location.
The supplementary video demonstrates the experiment.

\section{Conclusion and Future Work}
\label{sec:conclusion}
This paper presented an accurate and robust LiDAR-SLAM system on a resource constrained legged robot
using a factor graph-based optimization. We introduced an improved 
registration verification algorithm capable of running in real time. In addition, we leveraged a state-of-the-art learned loop 
closure detector which is sufficiently efficient to run online and had significant viewpoint robustness.
We examined our system in indoor and outdoor industrial environments with a final demonstration showing online operation of the system 
on our robot.

In the future, we will speed up our ICP registration to increase the update frequency in our SLAM system. 
We will also examine our system in more varied scenarios enabling real-time tasks on the quadruped 
ANYmal. In addition, we would like to integrate visual measurements taken from the robot's RGBD camera to
improve initialization of point cloud registration as well as independent registration verification.

\bibliographystyle{IEEEtran}
%\bibliography{library}

\end{document}